\documentclass[11pt,a4paper]{article}
\usepackage[hyperref]{naaclhlt2018}
\usepackage{times}
\usepackage{latexsym}
\usepackage{paralist}
\usepackage{multirow}
\usepackage{url}
\usepackage{graphicx}
\usepackage{amsmath, amssymb,verbatim}

\aclfinalcopy 
\def\aclpaperid{916} 

\title{Natural Language Generation by\\ Hierarchical Decoding with Linguistic Patterns}

\author{Shang-Yu Su$^\dagger$\quad Kai-Ling Lo$^\star$\quad Yi-Ting Yeh$^\star$\quad Yun-Nung Chen$^\star$ \\
$^\star$Department of Computer Science and Information Engineering\\
$^\dagger$Department of Electrical Engineering\\
National Taiwan University\\
\texttt{ \{f05921117,b04902010,b03902071\}@ntu.edu.tw\quad y.v.chen@ieee.org}
}

\date{}

\usepackage{lipsum}

\newcommand\blfootnote[1]{%
  \begingroup
  \renewcommand\thefootnote{}\footnote{#1}%
  \addtocounter{footnote}{-1}%
  \endgroup
}

\begin{document}
\maketitle
\begin{abstract}
Natural language generation (NLG) is a critical component in spoken dialogue systems. 
Classic NLG can be divided into two phases: (1) sentence planning: deciding on the overall sentence structure, (2) surface realization: determining specific word forms and flattening the sentence structure into a string. 
Many simple NLG models are based on recurrent neural networks (RNN) and sequence-to-sequence (seq2seq) model, which basically contains an encoder-decoder structure; these NLG models generate sentences from scratch by jointly optimizing sentence planning and surface realization using a simple cross entropy loss training criterion.
However, the simple encoder-decoder architecture usually suffers from generating complex and long sentences, because the decoder has to learn all grammar and diction knowledge.
This paper introduces a hierarchical decoding NLG model based on linguistic patterns in different levels, and shows that the proposed method outperforms the traditional one with a smaller model size.
Furthermore, the design of the hierarchical decoding is flexible and easily-extensible in various NLG systems\blfootnote{The first two authors have equal contributions.}\footnote{The source code is available at \url{https://github.com/MiuLab/HNLG}.}.

\end{abstract}

\section{Introduction}
\label{sec:intro}
Spoken dialogue systems that can help users to solve complex tasks have become an emerging research topic in artificial intelligence and natural language processing areas~\cite{wen2017network, bordes2017learning, dhingra2017towards, li2017end}. 
A typical dialogue system pipeline contains a speech recognizer, a natural language understanding component, a dialogue manager, and a natural language generator (NLG).

NLG is a critical component in a dialogue system, where its goal is to generate the natural language given the semantics provided by the dialogue manager.
As the endpoint of interacting with users, the quality of generated sentences is crucial for user experience. 
The common and mostly adopted method is the rule-based (or template-based) method~\cite{mirkovic2011dialogue}, which can ensure the natural language quality and fluency.
Considering that designing templates is time-consuming and the scalability issue, data-driven approaches have been investigated for open-domain NLG tasks.


Recent advances in recurrent neural network-based language model (RNNLM)~\cite{mikolov2010recurrent,mikolov2011extensions} have demonstrated the capability of modeling long-term dependency by leveraging RNN structure.
Previous work proposed an RNNLM-based NLG~\cite{wen2015stochastic} that can be trained on any corpus of dialogue act-utterance pairs without any semantic alignment and hand-crafted features. 
Sequence-to-sequence (seq2seq) generators~\cite{cho2014learning,sutskever2014sequence} further offer better results by leveraging encoder-decoder structure: previous model encoded syntax trees and dialogue acts into sequences~\cite{duvsek2016sequence} as inputs of attentional seq2seq model~\cite{bahdanau2014neural}.
However, it is challenging to generate long and complex sentences by the simple encoder-decoder structure due to grammar complexity and lack of diction knowledge.

\begin{figure*}[ht]
\centering
\includegraphics[width=\linewidth]{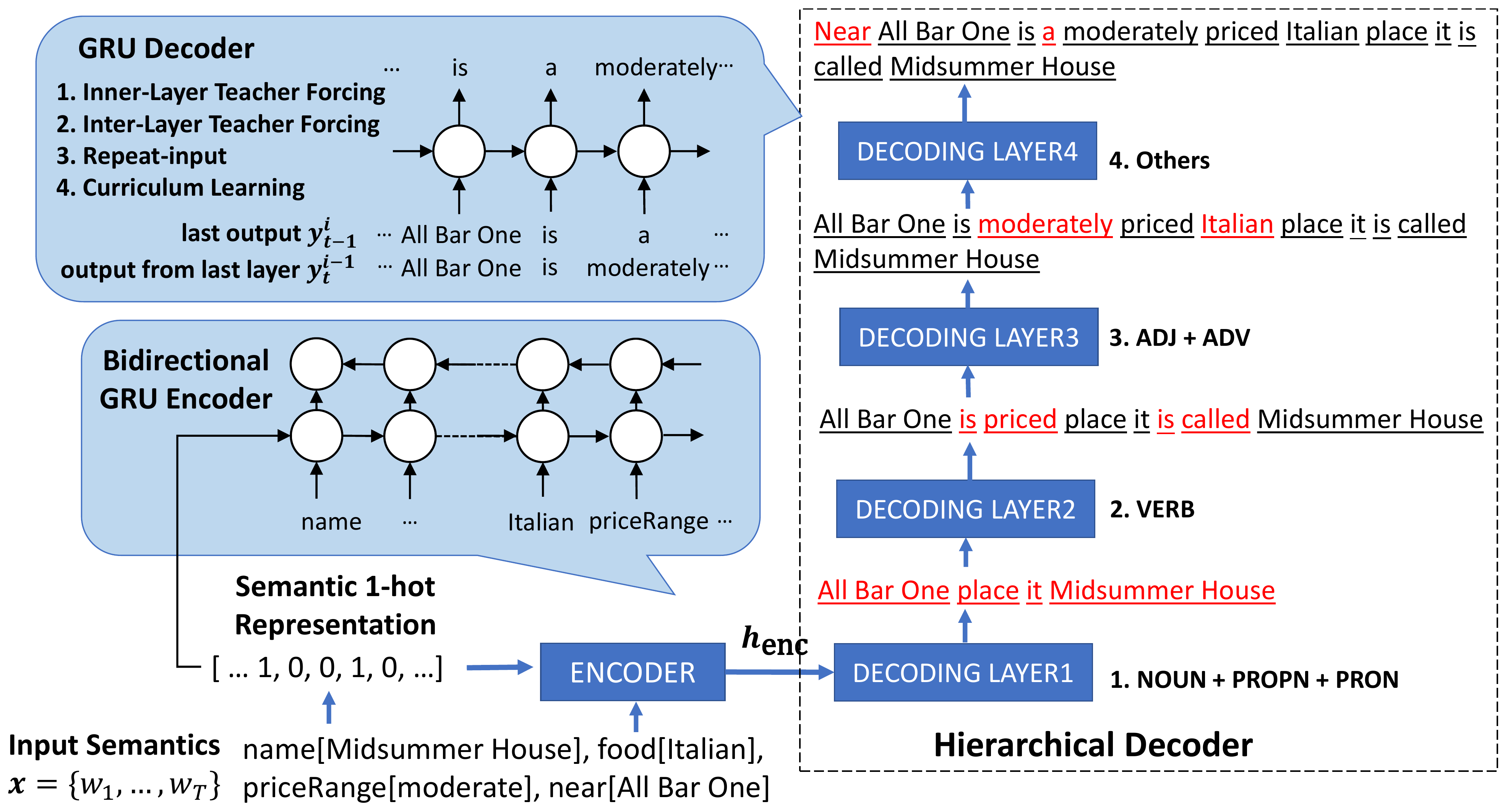}
\caption{The framework of the proposed semantically conditioned NLG model.}
\label{fig:framework}
\end{figure*}

This paper proposes a hierarchical decoder leveraging linguistic patterns, where the decoding hierarchy is constructed in terms of part-of-speech (POS) tags.
The original single decoding process is separated into a multi-level decoding hierarchy, where each decoding layer generates words associated with a specific POS set.
The experiments show that our proposed method outperforms the classic seq2seq model with  less parameters.
In addition, our proposed model allows other word-level or sentence-level characteristics to be further leveraged for better generalization.

\section{The Proposed Approach}
\label{sec:proposedapproach}

The framework of the proposed semantically conditioned NLG model is illustrated in Figure~\ref{fig:framework}, where the model architecture is based on an encoder-decoder (seq2seq) design~\cite{cho2014learning,sutskever2014sequence}.
In the seq2seq architecture, a typical generation process includes encoding and decoding phases: 
First, the given semantic representation sequence $\textbf{x}=\{w_t\}^T_1$ is fed into a RNN-based encoder to capture the temporal dependency and project the input to a latent feature space, 
and encoded into 1-hot semantic representation as the initial state of the encoder in order to maintain the temporal-independent condition as shown in the left-bottom of Figure~\ref{fig:framework}.
The recurrent unit of the encoder is bidirectional gated recurrent unit (GRU) ~\cite{cho2014learning}, 
\begin{equation}
\label{eq:basic}
\textbf{h}_\text{enc} = \text{BiGRU}(\textbf{x}).
\end{equation}
Then the encoded semantic vector, $\textbf{h}_\text{enc}$, flows into an RNN-based decoder as the initial state to generate word sequences by an RNN model shown in the left-top component of the figure.


\subsection{Hierarchical Decoder}
\label{ssec:hd}

Despite the intuitive and elegant design of the seq2seq model, it is difficult to generate long, complex, and decent sequences by such encoder-decoder structure, because a single decoder is not capable of learning all diction, grammar, and other related linguistic knowledge.
Some prior work applied additional technique such as reranker to select a better result among multiple generated sequences~\cite{wen2015stochastic,duvsek2016sequence}.
However, the issue still remains unsolved in NLG community.

Therefore, we propose a hierarchical decoder to address the above issue, where the core idea is to separate the decoding process and learn different types of patterns instead of learning all relevant knowledge together.
The hierarchical decoder is composed of several decoding layers, each of which is only responsible for learning a portion of the related knowledge.
Namely, the linguistic knowledge can be incorporated into the decoding process and divided into several subsets.

In this paper, we use part-of-speech (POS) tags as the additional linguistic features to construct the hierarchy, where POS tags of the words in the target sentence are separated into several subsets and each layer is responsible for decoding the words associated with a specific set of POS patterns.
An example is shown in the right part of Figure~\ref{fig:framework}, where the first layer at the bottom is in charge of learning to decode nouns, pronouns, and proper nouns, and the second layer is in charge of verbs, and so on. 
Our approach is also intuitive from the viewpoint of how humans learn to speak; for example, infants first learn to say the keywords which are often nouns.
When an infant says ``\emph{Daddy, toilet.}'', it actually means ``\emph{Daddy, I want to go to the toilet.}''. 
Along with the growth of the age, children learn more grammars and vocabulary and then start adding verbs to the sentences, further adding adverbs, and so on.
This process of how humans learn to speak is the core motivation of our proposed method.
 
In the hierarchical decoder,
the initial state of each GRU-based decoding layer $i$ is the extracted feature $\textbf{h}_\text{enc}$ from the encoder, and the input at every step is the last predicted token $\textbf{y}^i_{t-1}$ concatenated with the output from the previous layer $\textbf{y}^{i-1}_t$,
\begin{eqnarray}
\label{eq:basic}
\textbf{h}^i_t, \textbf{o}^i_t &=& \text{GRU}^i_\text{dec}(\textbf{y}^i_{t-1}, \textbf{y}^{i-1}_t \mid \textbf{h}_\text{enc}, \textbf{h}^i_{t-1}), \\
\textbf{y}^i_t &=& \texttt{argmax} (\textbf{o}_t),
\end{eqnarray}
where $\textbf{h}^i_t$ is the $t$-th hidden state of the $i$-th GRU decoding layer and $\textbf{y}^i_t$ is the $t$-th outputted word in the $i$-th layer.
The cross entropy loss is used for optimization.

\subsection{Inner- and Inter-Layer Teacher Forcing}
\label{ssec:interlayertf}
Teacher forcing~\cite{williams1989learning} is a strategy for training RNN that uses model output from a prior time step as an input, and it works by using the expected output at the current time step $\hat{\textbf{y}}_t$ as the input at the next time step, rather than the output generated by the network. 
In our proposed framework, an input of a decoder contains not only the output from the last step but one from the last decoding layer.
Therefore, we design two types of teacher forcing techniques -- inner-layer and inter-layer. 
\paragraph{Inner-layer teacher forcing} is the classic teacher forcing strategy:
\begin{eqnarray}
\label{eq:basic}
\textbf{h}^i_t, \textbf{o}^i_t = \text{GRU}^i_\text{dec}(\hat{\textbf{y}}^i_{t-1}, \textbf{y}^{i-1}_t \mid \textbf{h}_\text{enc}, \textbf{h}^i_{t-1}).
\end{eqnarray}
\paragraph{Inter-layer teacher forcing} uses the labels instead of the actual output tokens of the last layer:
\begin{eqnarray}
\label{eq:basic}
\textbf{h}^i_t, \textbf{o}^i_t = \text{GRU}^i_\text{dec}(\textbf{y}^i_{t-1}, \hat{\textbf{y}}^{i-1}_t \mid \textbf{h}_\text{enc}, \textbf{h}^i_{t-1}).
\end{eqnarray}
The teacher forcing techniques can also be triggered only with a certain probability, 
which is known as the scheduled sampling approach~\cite{bengio2015scheduled}. 
In our experiments, the scheduled sampling approach is also adopted.

\subsection{Repeat-Input Mechanism}
\label{ssec:rim}
The concept of our proposed method is to hierarchically generate the sequence, gradually adding words associated with different linguistic patterns.
Therefore, the generated sequences from the decoders become longer as the generating process proceeds to the higher decoding layers, and the sequence generated by a upper layer should contain the words predicted by the lower layers.
In order to ensure the output sequences with the constraints,
we design a strategy that repeats the outputs from the last layer as inputs until the current decoding layer outputs the same token, so-called repeat-input mechanism.
This approach offers at least two merits: 
(1) Repeating inputs tells the decoder that the repeated tokens are important to encourage the decoder to generate them. 
(2) If the expected output sequence of a layer is much shorter than the one of the next layer, the large difference in length becomes a critical issue of the hierarchical decoder, because the output sequence of a layer will be fed into the next layer. 
With the repeat-input mechanism, the impact of length difference can be mitigated.

\subsection{Curriculum Learning}
\label{ssec:cl}
The proposed hierarchical decoder consists of several decoding layers, the expected output sequences of upper layers are longer than the ones in the lower layers. 
The framework is suitable for applying the curriculum learning~\cite{elman1993learning}, in which core concept is that a curriculum of progressively harder tasks could significantly accelerate a network’s training. 
The training procedure is to train each decoding layer for some epochs from the bottommost layer to the topmost one.

\begin{table*}
\centering
\begin{tabular}{ | c| l | c c c c| }
    \hline
    \multicolumn{2}{|c|}{\bf NLG Model} & \bf \small BLEU & \bf \small ROUGE-1 & \bf \small ROUGE-2 & \bf \small ROUGE-L  \\
\hline \hline
(a) & Sequence-to-Sequence Model & 28.9 & 40.7 & 12.5 & 32.1 \\
(b) & + Hierarchical Decoder  & 43.1 & 53.0 & 24.6 & 40.4 \\
(c) & + Hierarchical Decoder, Repeat-Input & 42.3 & 52.9 & 24.0 & 40.1 \\
(d) & + Hierarchical Decoder, Curriculum Learning  & 58.4 & 60.4 & 30.6 & 44.6 \\
(e) & + All & \bf 58.7 & \bf 62.3 & \bf 31.6 & \bf 45.4 \\
    \hline
(f) & (e) with High Inner-Layer TF Prob. & \bf 62.1 & \bf 64.0 & \bf 32.8 & \bf 46.0 \\
(g) & (e) with High Inter-Layer TF Prob. & 56.7 & 61.3 & 30.9 & 44.6 \\
(h) & (e) with High Inner- and Inter-Layer TF Prob. & 60.0 & 63.0 & 31.8 & 45.2 \\
    \hline
  \end{tabular}
\caption{The NLG performance reported on BLEU, ROUGE-1, ROUGE-2, and ROUGE-L of models (\%).}
\label{tab:results}
\end{table*}


\section{Experiments}
\label{sec:exp}

\subsection{Setup}
\label{ssec:expsetup}
The experiments are conducted using the E2E NLG challenge dataset~\cite{novikova2017e2e}\footnote{\url{http://www.macs.hw.ac.uk/InteractionLab/E2E/}},
which is a crowd-sourced dataset of 50k instances in the restaurant domain. 
The input is the semantic frame containing specific slots and corresponding values, and the output is the natural language containing the given semantics as shown in Figure~\ref{fig:framework}.

To prepare the labels of each layer within the hierarchical structure of the proposed method, we utilize spaCy toolkit 
to perform POS tagging for the target word sequences. Some properties such as names of restaurants are delexicalized (for example, replaced with symbols "\texttt{RESTAURANT\_NAME}") to avoid data sparsity.
We assign the words with specific POS tags for each decoding layer: \textbf{nouns}, \textbf{proper nouns}, and \textbf{pronouns} for the first layer, \textbf{verbs} for the second layer, \textbf{adjectives} and \textbf{adverbs} for the third layer, and \textbf{others} for the forth layer.
Note that the hierarchies with more than four levels are also applicable, the proposed hierarchical decoder is a general and easily-extensible concept.

The experimental results are shown in Table~\ref{tab:results}, every reported number is averaged over the results on the official testing set from three different models.
Row (a) is the simple seq2seq model as the baseline.
The probability of activating inter-layer and inner-layer teacher forcing is set to 0.5 in the rows (a)-(e); to evaluate the impact of teacher forcing, the probability is set to 0.9 (rows (f)-(h)). The probability of teacher forcing is attenuated every epoch, and the decay ratio is 0.9.
We perform 20 training epochs without early stop; when the curriculum learning approach is applied, only the first layer is trained during first five epochs, the second decoder layer starts to be trained at the sixth epoch, and so on. 
To evaluate the quality of the generated sequences regarding both precision and recall, the evaluation metrics include BLEU and ROUGE (1, 2, L) scores with multiple references.

\subsection{Results and Analysis}
\label{ssec:resultandanalysis}
To fairly examine the effectiveness of our proposed approaches, we control the size of the proposed model to be smaller.
The baseline seq2seq decoder has 400-dim hidden layer, and the models with the proposed hierarchical decoder (rows (b)-(h)) have four 100-dim decoding layers.
Table~\ref{tab:results} shows that simply introducing the hierarchical decoding technique without adding parameters (row (b)) to separate the generation process into several phases achieves significant improvement in both BLEU and ROUGE scores, 49.1\% in BLEU, 30.2\% in ROUGE-1, 96.8\% in ROUGE-2, and 25.9\% in ROUGE-L. 
Applying the proposed repeat-input mechanism (row (c)) and the curriculum learning strategy (row (d)) both offer considerable improvement.
Combining all proposed techniques (row (e)) yields the best performance in both BLEU and ROUGE scores, achieving 103.1\%, 53.1\%, 152.8\%, and 41.4\% of relative improvement in BLEU, ROUGE-1, ROUGE-2, and ROUGE-L respectively.
The results demonstrate the effectiveness of the proposed approach.

To further verify the impact of teacher forcing, the integrated models (row (e)) with high inter and inner-layer teacher forcing probability (rows (f)-(h)) are also evaluated. 
Note that when the teacher forcing is activated probabilistically, the strategies are also known as schedule sampling~\cite{bengio2015scheduled}. 
Row (g) shows that high probability of triggering inter-layer teacher forcing results in slight performance degradation, while models with high inner-layer teacher forcing probability (rows (f) and (h)) can further benefit the model.

Note that the decoding process is a single-path forward generation without any heuristics and other mechanisms (like beam search and reranking), so the effectiveness of the proposed methods can be fairly verified. 
The experiments show that by considering linguistic patterns in hierarchical decoding, the proposed approaches can significantly improve NLG results with smaller models.

\section{Conclusion}
\label{sec:conclusion}
This paper proposes a seq2seq-based model with a hierarchical decoder that leverages various linguistic patterns and further designs several corresponding training and inference techniques.
The experimental results show that the models applying the proposed methods achieve significant improvement over the classic seq2seq model. 
By introducing additional word-level or sentence-level labels as features, the hierarchy of the decoder can be designed arbitrarily.
Namely, the proposed hierarchical decoding concept is general and easily-extensible, with flexibility of being applied to many NLG systems. 

\section*{Acknowledgements}
We would like to thank reviewers for their insightful comments on the paper.
The authors are supported by the Institute for Information Industry, Ministry of Science and Technology of Taiwan, Google Research, Microsoft Research,
and MediaTek Inc..

\bibliography{naaclhlt2018}
\bibliographystyle{acl_natbib}

\appendix

\section{Dataset Detail}
The experiments are conducted using the E2E NLG challenge dataset, which is a crowd-sourced dataset in the restaurant domain, the training set contains 42064 instances while there are 4673 instances in the validation (development) set. In our experiments, we use the validation set to test our models. 
In the E2E NLG Challenge dataset, the input is the semantics containing slots and their values, and the output is the corresponding natural language.
For example, the slot-value pairs \texttt{"name[Bibimbap House], food[English], priceRange[moderate], area[riverside], near[Clare Hall]"}  correspond to the target sentence ``\emph{Bibimbap House is a moderately priced restaurant who's main cuisine is English food. You will find this local gem near Clare Hall in the Riverside area.}''.

\section{Parameter Setting}
We use mini-batch Adam as the optimizer with the batch size of 32 examples.
The baseline seq2seq model (row (a)) sets the encoder's hidden layer size to 200 and the decoder's to 400.
The size of the hidden layer in the encoder and the decoder layers of the models based on the proposed hierarchical decoder (rows (b)-(h)) are 200 and 100, respectively.
Note that in this setting, the models applied the proposed methods will have less parameters than the baseline seq2seq model. In terms of the models utilized the basic RNN cell, the baseline seq2seq model (row (a)) has 640k parameters whereas the proposed models (rows (b)-(h)) have only 520k parameters.

\end{document}